# Unsupervised Learning of Noisy-Or Bayesian Networks

**Yoni Halpern, David Sontag**
Department of Computer Science
Courant Institute of Mathematical Sciences
New York University

## Abstract

This paper considers the problem of learning the parameters in Bayesian networks of discrete variables with known structure and hidden variables. Previous approaches in these settings typically use expectation maximization; when the network has high treewidth, the required expectations might be approximated using Monte Carlo or variational methods. We show how to avoid inference altogether during learning by giving a polynomial-time algorithm based on the method-of-moments, building upon recent work on learning discrete-valued mixture models. In particular, we show how to learn the parameters for a family of bipartite noisy-or Bayesian networks. In our experimental results, we demonstrate an application of our algorithm to learning QMR-DT, a large Bayesian network used for medical diagnosis. We show that it is possible to fully learn the parameters of QMR-DT even when only the findings are observed in the training data (ground truth diseases unknown).

## 1 Introduction

We address the problem of unsupervised learning of the parameters of bipartite noisy-or Bayesian networks. Networks of this form are frequently used models for expert systems and include the well-known Quick Medical Reference (QMR-DT) model for medical diagnosis (Miller *et al.*, 1982; Shwe *et al.*, 1991).

Given that QMR-DT is one of the most well-studied noisy-or Bayesian networks, we use it as a running example for the type of network that we would like to provably learn. It is a large bipartite network, describing the relationships between 570 binary disease variables and 4,075 binary symptom variables using 45,470 directed edges. It was laboriously assembled based on information elicited from experts and represents an example of a network that captures (at least some of) the complexities of real-world medical diagnosis tasks.

Learning these parameters is important. Both the structure and the parameters of the QMR-DT model were manually specified, taking over 15 person-years of work (Miller *et al.*, 1982). Each disease took one to two weeks of full-time effort, involving in-depth review of the medical literature, to incorporate into the model. Despite this effort, the original INTERNIST-1/QMR model still lacked an estimated 180 diseases relevant to general internists (Miller *et al.*, 1986). Furthermore, model parameters such as the priors over the diseases can vary over time and location.

Although it is often possible to extract symptoms or findings from unstructured clinical data, obtaining reliable ground truth for a patient's underlying disease state is much more difficult. Often all we have available are noisy and biased estimates of the patient's disease state in the form of billing or diagnosis codes and free text. We can, however, treat these noisy labels as additional findings (for training) and perform unsupervised learning. The ability to learn parameters from unlabeled data could make models like QMR-DT much more widely applicable.

Exact inference in the QMR-DT network is known to be intractable (Cooper, 1987), so it would be expected to resort to expectation-maximization techniques using approximate inference in order to learn the parameters of the model (Jaakkola & Jordan, 1999; Šingliar & Hauskrecht, 2006). However, these methods can be computationally costly and are not guaranteed to recover the true parameters of the network even when presented with infinite data drawn from the model.

We give a polynomial-time algorithm for provably learning a large family of bipartite noisy-or Bayesian networks. It is important to note that this method does not extend to all bipartite networks. It does not work on certain densely connected structures. We provide a criterion based on the network structure to determine whether or not the network is learnable by our algorithm. Though the algorithm is limited, the family of networks for which we can learn parameters is certainly non-trivial.

Our approach is based on the method-of-moments, and builds upon recent work on learning discrete-valued mixture models (Anandkumar *et al.* , 2012c; Chang, 1996; Mossel & Roch, 2005). We assume that the observed data is drawn independently and identically distributed from a model of known structure and unknown parameters, and show that we can accurately and efficiently recover those parameters with high probability using a reasonable number of samples. Making these additional assumptions allows us to circumvent the hardness of maximum likelihood learning.

Our parameter learning algorithm begins by finding triplets of observed variables that are *singly-coupled*, meaning that they are marginally mixture models. After learning the parameters involving these, we show how one can *subtract* their influence from the empirical distribution, which then allows for more parameters to be learned. This process continues until no new parameters can be learned. Surprisingly, we show that this simple algorithm is able to learn almost all of the parameters of the QMR-DT structure. Finally, we study the identifiability of the learning problem with hidden variables and show that even in dense networks, the true model is often identifiable from third-order moments. Our identifiability results suggest that the final parameters of QMR-DT can be learned with a grid search over a single parameter.

We see the significance of our work as presenting one of the first polynomial-time algorithms for learning a family of discrete-valued Bayesian networks with hidden variables where exact inference on the hidden variables is intractable. We believe that our algorithm will be of practical interest in applications (such as medical diagnosis) where prior knowledge can be used to specify the Bayesian network structure involving the hidden variables and the observed variables.

## 2 Background

We consider bipartite noisy-or Bayesian networks with $n$ binary latent variables, $D = \{D_1, D_2, ..., D_n\}, D_i \in \{0,1\}$, and $m$ observed binary variables, $S = \{S_1, S_2, ..., S_m\}, S_i \in \{0,1\}$. Continuing with the medical diagnosis example, we refer to the latent variables as *diseases* and the observed variables as *symptoms*. The edges in the model are directed from the latent diseases to the observed symptoms. We assume that the diseases are never observed, neither at training nor test time, and show how to recover the parameters of the model in an unsupervised manner.

By using a noisy-or conditional distribution to model the interactions from the latent variables to the observed variables, the entire Bayesian network can be parametrized by $n \times m + n + m$ parameters. These parameters consist of *prior* probabilities on the diseases $\Pi = \{p_1, p_2, ..., p_n\}$, *failure* probabilities between diseases and symptoms, $F = \{\vec{f_1}, \vec{f_2}, ... \vec{f_n}\}$, where each $\vec{f_i}$ is a vector of size $m$, and noise (or leak) probabilities $\vec{\nu} = \{\nu_1, ... \nu_m\}$. An equivalent formulation includes the noise in the model by introducing a single 'noise' disease, $d_0$, which is present with probability $p_0 = 1$ and has failure probabilities $\vec{f_0} = 1 - \vec{\nu}$.

Observations are sampled from the noisy-or network by the following generative process:
— The set of present diseases is drawn according to Bernoulli($\Pi$).
— For each present disease $D_i$, the set of active edges $\vec{a_i}$, is drawn according to Bernoulli($1 - \vec{f_i}$).
— The observed value of the $j^{th}$ symptom is then given by $s_j = \bigcup_i a_{i,j}$ (this part is deterministic).

While the network can be described generally as being fully connected, in practice many of the diseases have zero probability of generating many of the symptoms (ie. fail with probability 1). The Bayesian network only has an edge between disease $D_i$ and symptom $S_j$ if $f_{i,j} < 1$. As we explain in Section 3, our ability to learn parameters will depend on the particular sparsity pattern of these edges.

The marginal distribution over a set of symptoms, $\mathcal{S}$, in the noisy-or network has the following form:

$$p(\mathcal{S}) = \sum_{\{D\}} \prod_{i=1}^{n} p(d_i) \prod_{j \in \mathcal{S}} p(s_j|D), \quad (1)$$

where $\{D\}$ is the set of $2^n$ configurations of the disease variables $\{d_1, ..., d_n\}$. The disease priors are given by $p(d_i) = p_i^{d_i}(1-p_i)^{1-d_i}$, and the conditional distribution of the symptoms by a noisy-or distribution:

$$p(s_j|D) = \left(1 - f_{0,j} \prod_{i=1}^{n} f_{i,j}^{d_i}\right)^{s_j} \left(f_{0,j} \prod_{i=1}^{n} f_{i,j}^{d_i}\right)^{1-s_j} \quad (2)$$

The algorithms described in this paper make substantial use of sets of moments of the observed variables. The first moment that will be important is the joint distribution over a set of symptoms, $\mathcal{S}$, which we call $T_\mathcal{S}$. $T_\mathcal{S}$ is a $|\mathcal{S}|^{th}$ order tensor where each dimension is of size 2. For a set of symptoms $\mathcal{S} = (S_a, S_b, S_c)$ the elements of $T_\mathcal{S}$ are defined as: $T_{\mathcal{S}(s_a, s_b, s_c)} = p(S_a = s_a, S_b = s_b, S_c = s_c)$. Throughout the paper we will make use of sets of at most three variables, so the joint distributions are of maximal size $2 \times 2 \times 2$.

We also make use of the negative moment of a set of symptoms $\mathcal{S}$, which we denote as $\bar{M}_\mathcal{S}$, defined as the marginal probability of observing *all* of the symptoms in $\mathcal{S}$ to be absent. The negative moments of $\mathcal{S}$ have the following compact form:

$$\bar{M}_\mathcal{S} \equiv p(\bigcap_{S_j \in \mathcal{S}} S_j = 0) = \prod_{i=0}^{n} \left(1 - p_i + p_i \prod_{S_j \in \mathcal{S}} f_{i,j}\right) \quad (3)$$

The form of Eq. 3 makes it clear that the parameters associated with each parent are all grouped together in a single term, which we call the *influence* of disease $D_i$ on symptoms $\mathcal{S}$. Define this influence term to be $I_{i,\mathcal{S}} \equiv 1 - p_i + p_i \prod_{S_j \in \mathcal{S}} f_{i,j}$. Using this, we rewrite Eq. 3 using influences as $\bar{M}_\mathcal{S} = \prod_{i=0}^{n} I_{i,\mathcal{S}}$. This formulation is found in Heckerman (1990) and provides a compact form that makes it easy to take advantage of the noisy-or properties of the network.

### 2.1 Related Work

The problem of inference in bipartite noisy-or networks with fixed parameters has been studied and exact inference in large models like the QMR-DT model is known to be intractable (Cooper, 1987). The Quickscore formulation by Heckerman (1990) takes advantage of the noisy-or parameterization to give an exact inference algorithm that is polynomial in the number of negative findings but still exponential in the number of positive findings.

Any expectation maximization (EM) approach to learning the network parameters must contend with the computational complexity of inference in these models. Many approximate inference strategies have been developed, notably Jaakkola & Jordan (1999) and Ng & Jordan (2000). The closest related work to our paper is by Šingliar & Hauskrecht (2006), who give a variational EM algorithm for unsupervised learning of the parameters of a noisy-or network. We will use their algorithm as a baseline in our experimental results. Importantly, variational EM algorithms do not have consistency guarantees.

Kearns & Mansour (1998) develop an inference-free approach which is guaranteed to learn the exact structure and parameters of a noisy-or network under specific identifiability assumptions by performing a search over network structures. In order to achieve their results, they impose strong constraints such as identical priors on all of the parents. Their structure learning algorithm is exponential in the maximal in-degree of the symptom nodes, which for QMR-DT is 570. More importantly, the overall approach relies on the model family having a property called "unique polynomials", closely related to the question of identifiability, but which is left mostly uncharacterized in their paper. It is not clear whether their algorithm can be modified to take advantage of a known structure. As such, no existing method is sufficient for learning the parameters of a large network like the QMR-DT network.

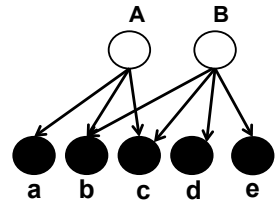

Figure 1: A small noisy-or network. The triplets (b,d,e) and (c,d,e) are both singly-coupled by $B$. The presence of disease $B$ prevents (a,b,c) from being singly-coupled. However, after learning the parameters of disease $B$, we can subtract off its influence, leaving (a,b,c) singly-coupled.

Spectral approaches to learning mixture models originated with Chang's spectral method (Chang 1996; analyzed in Mossel & Roch 2005). These methods have been successfully applied to learning discrete mixture models and hidden Markov models (Anandkumar et al. , 2012c), as well as continuous admixture models such as latent Dirichlet allocation (Anandkumar et al. , 2012b). In recent work, these have been generalized to a large class of linear latent variable models (Anandkumar et al. , 2012d). However, the noisy-or model is not linear, making it non-trivial to apply these methods that rely on linearity of expectation to relate the general formula for observed moments to a low rank matrix or tensor decomposition.

## 3 Parameter Learning with Known Structure

In this section we present a learning algorithm that takes advantage of the known structure of a noisy-or network in order to learn parameters using only low-order moments. We first identify singly-coupled triplets, which are marginally mixture models and therefore we can learn their parameters. Once some parameters of the network are learned, we make adjustments to the observed moments, subtracting off the influence of some parents, essentially removing them from the network, making more triplets singly-coupled (illustrated in Figure 1). Algorithm 1 outlines the parameter learning procedure.

We discuss the running time in Section 3.4. The clean up procedure is not part of the main algorithm and may increase the runtime to exponential, depending on the configuration of the network of remaining parameters at the end of the main algorithm. We present it because it allows us to extend the algorithm to learn the QMR-DT network which has a very simple network of remaining parameters after running the main algorithm to completion.

**Algorithm 1** Learn Parameters
---
Inputs: A bipartite noisy-or network structure with unknown parameters $F, \Pi$. $N$ samples from the network.
Outputs: Estimates of $F$ and $\Pi$.
– *Main Routine*
 1: unknown = $\{f_{i,j} \in F\} \cup \{p_i \in \Pi\}$
 2: knowns = $\{\}$
 3: **while** not converged **do**
 4:  learned = $\{\}$
 5:  **for all** $f_{i,a}$ in unknown_parameters **do**
 6:   **for all** $(S_b, S_c)$, siblings of $S_a$ **do**
 7:    Parents = parents of $(S_a, S_b, S_c)$
 8:    knownParents = All $D_k$ in Parents for which $f_{k,a}, f_{k,b}$ and $f_{k,c}$ are known.
 9:    Remove knownParents from the graph.
 10:    **if** $(S_a, S_b, S_c)$ are singly-coupled (Def. 1) **then**
 11:     Form joint distribution $T_{a,b,c}$
 12:     **for all** $D_k$ in knownParents **do**
 13:      $T_{a,b,c}$ = RemoveInfluence($T_{a,b,c}$, $D_k$) (Section 3.2)
 14:     **end for**
 15:     Learn $p_i, f_{i,a}, f_{i,b}, f_{i,c}$. (Eq. 4)
 16:     unknown = unknown - $(p_i, f_{i,a}, f_{i,b}, f_{i,c})$.
 17:     learned = learned $\cup (p_i, f_{i,a}, f_{i,b}, f_{i,c})$
 18:    **end if**
 19:    Add back knownParents to the graph.
 20:   **end for**
 21:  **end for**
 22:  known = known $\cup$ learned
 23:  Converge if no new parameters are learned.
 24: **end while**
 25: Learn noise parameters (Eq. 5).
– *Clean up*
 1: Check identifiability of remaining parameters with third-order moments and use clean up procedure to learn remaining parameters. (Section 3.3)
---

The algorithm can be further optimized by precomputing and storing dependencies between triplets (i.e., triplet $A$ can be learned after triplet $B$ is learned) to avoid repeated searches for singly-coupled triplets. The algorithm is also greedy in that it learns each failure parameter $f_{i,j}$ with the first suitable triplet it encounters. A more sophisticated version would attempt to determine the best triplet to learn $f_{i,j}$ with high confidence, which we do not explore in this paper.

The following sections go into more detail on the various steps of the algorithm, and assume that we have access to the exact moments (i.e., infinite data). In Section 3.4 we show that the error incurred by using sample estimates of the expectations is bounded.

### 3.1 Learning Singly-coupled Symptoms

The condition that we require to learn the parameters is that the observed variables be *singly-coupled*:

**Definition 1.** *A set of symptoms, $\mathcal{S}$ is* singly-coupled *by parent $D_i$ if $D_i$ is a parent of $S_j$ for all $S_j \in \mathcal{S}$ and there is no other parent, $D_k \in \{D_1, ..., D_n\}$, such that $D_k$ is a parent of at least two symptoms in $\mathcal{S}$.*

The intuition behind using singly coupled symptoms is they can be viewed locally as mixture models with two mixture components corresponding to the states of the coupling parent. For example, in Figure 1, $(b, d, e)$ and $(c, d, e)$ form singly-coupled triplets coupled by disease $B$. Their observations are independent conditioned on the state of $B$. The noise disease, $D_0$, does not have to be considered here since it is present with probability 1, and so its state is always observed. Thus, the noise parent can never act as a coupling parent.

Observing that the singly-coupled condition locally creates a binary mixture model, we conclude that we can learn the noisy-or parameters associated with a singly-coupled triplet by using already existing methods for learning 3-view mixture models from the third-order moment $T_{a,b,c}$. While the general method of learning multi-view mixture models described in Anandkumar *et al.* (2012a) would suffice, we employ a simpler method (given in Algorithm 2) applicable to mixture models of binary variables based on a tensor decomposition described in Berge (1991). This procedure uniquely decomposes $T_{a,b,c}$ into two rank-1 tensors which describe the conditional distributions of the symptoms conditioned on the state of the parent.

The tensor decomposition returns the prior probabilities of the parent states and the probabilities of the children conditioned on the state of the parent. Ambiguity in the labeling of the parent states is avoided since for noisy-or networks $p(S_j = 0 | D_i = 0) > p(S_j = 0 | D_i = 1)$. To obtain the noisy-or parameters, we observe that the prior for the disease is simply given by the mixture prior, and the failure probability $f_{i,j}$ between the coupling disease $D_i$ and symptom $S_j$ is the ratio of two conditional probabilities:

$$p_i = p(D_i = 1), \quad f_{i,j} = \frac{p(S_j = 0 | D_i = 1)}{p(S_j = 0 | D_i = 0)}. \quad (4)$$

The noise parameter $f_{0,j}$ is not learned using the above equations since $D_0$ never acts as a coupling parent. However, once all of the other parameters are learned, the noise parameter simply provides for any otherwise

**Algorithm 2** Binary Tensor Decomposition

Input: Tensor $T$ of size $2 \times 2 \times 2$ which is a joint probability distribution over three variables $(S_a, S_b, S_c)$ which are singly-coupled by disease $Z$.
Output: Prior probability $p(Z = 1)$, and conditional distributions $p(s_a, s_b, s_c | Z = 0)$, $p(s_a, s_b, s_c | Z = 1)$.

1: Matrix $X_1 = T_{(0,\cdot,\cdot)}$
2: Matrix $X_2 = T_{(1,\cdot,\cdot)}$
3: $Y_2 = X_2 X_1^{-1}$
4: *Find eigenvalues of $Y_2$ using quadratic equation:*
5: $\lambda_1, \lambda_2 = \text{roots}(\lambda^2 - \text{Tr}(Y_2)\lambda + \text{Det}(Y_2))$
6: $\vec{u}_1 \vec{v}_1^T = (\lambda_1 - \lambda_2)^{-1}(X_2 - \lambda_2 X_1)$
7: $\vec{u}_2 \vec{v}_2^T = -(\lambda_1 - \lambda_2)^{-1}(X_2 - \lambda_1 X_1)$
8: Decompose* $\vec{u}_1 \vec{v}_1^T$, $\vec{u}_2 \vec{v}_2^T$ into $\vec{u}_1, \vec{u}_2, \vec{v}_1, \vec{v}_2$.
9: $\vec{l}_1 = \begin{pmatrix} 1 & \lambda_1 \end{pmatrix}^T$, $\vec{l}_2 = \begin{pmatrix} 1 & \lambda_2 \end{pmatrix}^T$
10: $T_1 = \vec{u}_1 \otimes \vec{v}_1 \otimes \vec{l}_1$, $T_2 = \vec{u}_2 \otimes \vec{v}_2 \otimes \vec{l}_2$
11: **if** $T_{1(0,0,0)} > T_{2(0,0,0)}$ **then**
12:     swap $T_1, T_2$
13: **end if**
14: $p(Z = 1) = \sum_{i,j,k} T_{2(i,j,k)}$
15: normalize $p(s_a, s_b, s_c | Z = 0) = T_1 / \sum_{i,j,k} T_{1(i,j,k)}$
16: normalize $p(s_a, s_b, s_c | Z = 1) = T_2 / \sum_{i,j,k} T_{2(i,j,k)}$

*To decompose the $2 \times 2$ matrix $\vec{u}\vec{v}^T$ into vectors $\vec{u}$ and $\vec{v}$, set $\vec{v}^T$ to the top row and $\vec{u}^T = \begin{pmatrix} 1 & \frac{(\vec{u}\vec{v}^T)_{(2,2)}}{(\vec{u}\vec{v}^T)_{(1,2)}} \end{pmatrix}$.

–Notation $T = \vec{u} \otimes \vec{v} \otimes \vec{l}$ means that $T_{(i,j,k)} = u_i v_j l_k$.

unaccounted observations, i.e.

$$f_{0,j} = \frac{\bar{M}_j}{\prod_{D_i \in Parents(S_j)} I_{i,j}}. \quad (5)$$

### 3.2 Adjusting Moments

Consider a triplet $(a, b, c)$ which has a common parent $A$, but is not singly coupled due to the presence of a parent $B$ shared by $b$ and $c$ (Figure 1). If we wish to learn the parameters involving this triplet and $A$ using the methods described above, we would need to form an adjusted moment, $\tilde{T}_{a,b,c}$ which would describe the joint distribution of $(a, b, c)$ if $B$ did not exist.

The influence of $B$ on variables $(b, c)$ is fully described by the parameters $p_B, f_{B,b}, f_{B,c}$. Thus, if we have estimates for these parameters, we can remove the influence of B to form the joint distribution over $(a, b, c)$ as though $B$ did not exist. This can be seen explicitly in Equation 3. In this form, the influence of each parent, if known, can be isolated and removed from the negative moments with a division operation. Since all the variables are binary, the mapping between the negative moments and the joint distribution is simple and the adjusted joint distribution can be formed from the power set of adjusted negative moments.

This procedure of adjusting moments by removing the influence of parents vastly expands the class of networks whose parameters are fully learnable using the singly-coupled triplet method from Section 3.1. Using these methods, complicated real-world networks such as the QMR-DT network can be learned almost fully. The clean up procedure described in the next section will make it possible to learn the remaining parameters of the QMR-DT network.

### 3.3 Extensions of the Main Algorithm

**Learning with singly-coupled pairs**. It is not possible to identify the parameters of a noisy-or model by only looking at singly-coupled pairs. However, once we have information about some of the parameters from looking at triplets, we can use it to find more parameter values by examining pairs. For example, in Figure 1, if $p_B$ and $f_{B,d}$ were learned using the triplet $(b, d, e)$, it would be possible to find $f_{B,c}$ using only the pairwise moment between $(c, d)$. More generally, for a singly-coupled pair of observables $(S_i, S_j)$ coupled by parent $D_i$, the following linear equation holds and can be used to solve for the unknown $f_{i,k}$ assuming $f_{i,j}$ and $p_i$ are already estimated:

$$\frac{\bar{M}_{\{j,k\}}}{\bar{M}_j \bar{M}_k} = \frac{1 - p_i + p_i f_{i,j} f_{i,k}}{(1 - p_i + p_i f_{i,j})(1 - p_i + p_i f_{i,k})}. \quad (6)$$

Thus, once some parameters have been estimated, singly-coupled pairs provide an alternative to singly-coupled triplets. Extending Algorithm 1 to search for singly-coupled pairs as well as triplets is trivial. For complex networks, using pairs allows us to learn more parameters with fewer adjustment steps.

**Clean up procedure**. For some Bayesian network structures, after running the main algorithm to completion, we may be left with some unlearned parameters. This occurs because it may be impossible to find enough singly-coupled triplets and pairs.

In these settings, it is natural to ask whether it is possible to uniquely identify the remaining parameters. We use a technique developed by Hsu *et al.* (2012) to show that most fully connected bipartite networks are *locally identifiable*, meaning that they are identifiable on all but a measure zero set of parameter settings. In particular, we use their CheckIdentifiability routine, which computes the Jacobian matrix of the system of moment constraint equations and evaluates its rank at a random setting of the parameters. We start with first-order moments and increase the order until the Jacobian is full rank, which implies that the model is locally identifiable with these moments. When the test succeeds it gives hope that, for all but a very small number of pathological cases, the networks can still be identifiable (up to a trivial relabeling of parents).

| | Number of Symptoms | | | | | | |
|---|---|---|---|---|---|---|---|
| Number of Hidden Variables | | 1 | 2 | 3 | 4 | 5 | 6 | 7 |
| 1 | -1 | -1 | 3 | 3 | 3 | 3 | 3 |
| 2 | -1 | -1 | -1 | 3 | 3 | 3 | 3 |
| 3 | -1 | -1 | -1 | -1 | 3 | 3 | 3 |
| 4 | -1 | -1 | -1 | -1 | 4 | 3 | 3 |
| 5 | -1 | -1 | -1 | -1 | -1 | 3 | 3 |
| 6 | -1 | -1 | -1 | -1 | -1 | 4 | 3 |
| 7 | -1 | -1 | -1 | -1 | -1 | 4 | 3 |

Table 1: Identifiability of parameters in fully-connected bipartite networks. Each row represents a number of hidden variables and each column is the number of observed variables. The value at location $(i, j)$ is the number of moments required to make the model identifiable according to the local identifiability criteria of the Jacobian method. E.g., 3rd order moments are needed to learn with a single hidden variable. The value -1 means the model is not identifiable even with the highest possible order moments.

Table 1 summarizes the results on networks with varying number of children. Even for fully connected networks, third-order moments are sufficient to satisfy the local identifiability criteria provided that there are a sufficient number of children.[1]

At this point, we can make progress by relying on the identifiability of the network from third-order moments and doing a grid search over parameter values to find the values that best match the observed third-order moments. For example, consider the network in Figure 2. This could be a sub-network that is left to learn after a number of other parameters have been learned and possibly removed. If we knew the values for the prior $p_A$ and failure probability $f_{A,a}$, then we would learn all of the edges from $A$ and subtract them off using the pairs learning procedure. When we do not know $p_A$ and $f_{A,a}$, we can search over the range of values and choose the values that yield the closest third-order moments to the observed moments.

Significantly, this method of doing a grid search over two parameters can be used no matter how many children are shared by $A$ and $B$. It only depends on the number of parents whose parameters are not learned. Thus, even if there are a large number of parameters left at the end of the main algorithm, we can proceed efficiently if they belong to a small number of parents. In Section 4.2 we note that in the QMR-DT network, all of the parameters that are left at the end of the main algorithm belong to only two parents and thus can be learned efficiently using the clean up phase.

---

[1]Third-order moments are also necessary for identifiability. Appendix G of Anandkumar *et al.* (2012a) gives an example of two networks, each with a single latent variable and three observations, that are indistinguishable using only second-order moments.

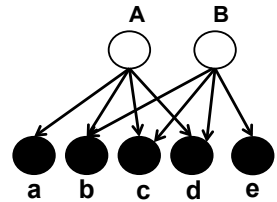

Figure 2: Similar to Figure 1, with the addition of a single edge from $A$ to $d$. There are now no singly-coupled triplets and learning cannot proceed. In the clean up procedure, we perform a grid search over values for $p_A$ and $f_{A,a}$, use them to learn all of the edges leading to $A$ and then proceed to subtract off the influence of $A$ and learn the edges of $B$.

### 3.4 Theoretical Properties

**Valid schedule.** We call a schedule, describing an order of adjustment and learning steps, *valid* if every learning step operates on a singly-coupled triplet (possibly after adjustment) and every parameter used in an adjustment is learned in a preceding step.

Note that a schedule is completely data independent, and depends only on the structure of the network. Algorithm 1 can be used to find a valid schedule if one exists. A valid schedule can also be used as a certificate of parameter identifiability for noisy-or networks with known structure:

**Theorem 1.** *If there exists a valid schedule for a noisy-or network, then all parameters are uniquely identifiable using only third-order moments.*

The proof follows from the uniqueness of the tensor decomposition described in Berge (1991).

**Computational complexity.** We run Algorithm 1 in two passes. In the first pass, we take as input the structure and find a valid schedule. The schedule will use one triplet per edge $f_{ij} \in F$, resulting in at most $|F|$ triplets for which to estimate the moments. Next, we iterate through the data, computing the required statistics. Finally, we do a second pass with the schedule to learn the parameters. The running time without the clean up procedure is $O(nm^2|F|^2 + |F|N)$, where $N$ is the number of samples.

**Sample complexity.** The parameter learning and adjustments presented above recover the parameters of the network exactly under the assumption that perfect estimates of the moments are available. With finite data sampled i.i.d. from a noisy-or network, the estimates of the moments are subject to sampling noise. In what follows, we bound the error accumulation due to using imperfect estimates of the moments.

Since error accumulates with each learning and ad-justment step, we define the *depth* of a parameter $\theta$ to be the number of extraction and adjustment steps required to reach the state in which $\theta$ can be learned. This depth is defined recursively:

**Definition 2.** *Denote the parameters used in the adjustment step before learning $\theta$ as $\Theta_{adj}$. $Depth(\theta) = \max_{\theta_i \in \Theta_{adj}} Depth(\theta_i) + 1$. If no adjustment is needed to learn $\theta$ then we say its depth is 0.*

To ensure that parameters are learned with the minimum depth, we construct the schedule in rounds. In round $k$ we learn all parameters that can be learned using parameters learned in previous rounds. We only update the set of known parameters at the end of the round. In this manner we are ensured that at each round, the algorithm learns all of the parameters that can be learned at a given depth.

The sample complexity result will depend on how close the parameters of the model are to 0 or 1. In particular, we define $p_{\min}$, $p_{\max}$ as the minimum and maximum disease priors, and $f_{\max}$ as the maximum failure probability. Additionally, we define $\bar{M}_{\min} = \min_{S_j \in S} \Pr(S_j = 0)$ to be the minimum marginal probability of any symptom being absent.

Our algorithm makes black-box use of an algorithm for learning mixture models of binary variables. In giving our sample complexity result, we abstract the dependence on the particular mixture model learning algorithm as follows:

**Definition 3.** *Let $f(\bar{M}_{\min}, f_{max}, p_{\max}, p_{\min}, \hat{\delta})$ be a function that represents the multiplicative increase in error incurred by learning the parameters of a mixture model from an estimate $\hat{T}_{a,b,c}$ of the third-order moment $T_{a,b,c}$, such that for all mixture parameters $\theta$,*

$$||\hat{T}_{a,b,c} - T_{a,b,c}||_1 < \hat{\epsilon} \implies$$
$$|\hat{\theta} - \theta| < f(\bar{M}_{\min}, f_{max}, p_{\max}, p_{\min}, \hat{\delta})\hat{\epsilon}$$

*with probability at least $1 - \hat{\delta}$.*

Using this, we obtain the sample complexity result ($K$ refers to the maximal in-degree of any symptom):

**Theorem 2.** *Let $\Theta$ be the set of parameters to be learned. Given a noisy-or network with known structure and a valid schedule with some constant maximal depth $d$, after a number of samples equal to*

$$N = \tilde{O}\Big(\Big(f\Big(\bar{M}_{\min}, f_{max}, p_{\max}, p_{\min}, \frac{\delta}{|\Theta|K^d}\Big)\Big)^{2d+2} \cdot$$
$$K^{2d}\bar{M}_{\min}^{-6d} \cdot \epsilon^{-2} \cdot \ln(|\Theta|/\delta)\Big)$$

*and with probability $1 - \delta$, for all $\theta \in \Theta$ Algorithm 1 returns an estimate $\hat{\theta}$ such that $|\hat{\theta} - \theta| < \epsilon$. This holds for $\epsilon < \frac{1}{2} f\Big(\bar{M}_{\min}, f_{max}, p_{\max}, p_{\min}, \frac{\delta}{|\Theta|K^d}\Big)^{-1} \Big(\frac{\bar{M}_{\min}^3}{15K}\Big)$.*

The proof consists of bounding the error incurred at each successive operation of learning parameters, using them to adjust the joint distributions, and applying standard sampling error bounds. The multiplicative increase in error with every adjustment and learning step leads to an exponential increase in error when these steps are applied repeatedly in series. The dependence on the maximal in-degree, $K$, comes from the possibility that in any adjustment step it may be necessary to subtract off the influence of all but one parent of the symptoms in the triplet. The maximum value for $\epsilon$ comes from division operations in both the learning and adjustment steps. If $\epsilon$ is not sufficiently small then the error can blow up in these steps.

Using the bounds presented for the mixture model learning approach in Anandkumar *et al.* (2012a) gives

$$f(\bar{M}_{\min}, f_{max}, p_{\max}, p_{\min}, \hat{\delta}) \propto \bar{M}_{min}^{-11}(1 - f_{max})^{-10}$$
$$\cdot (\min\{1 - p_{max}, p_{min}\})^{-2} \cdot \frac{\ln(1/\hat{\delta})}{\hat{\delta}},$$

though these bounds may not be tight. In particularly, the $\frac{1}{\delta}$ dependency in $f$ comes from a randomized step of the learning procedure. For binary variables this step may not be necessary and the $\frac{1}{\delta}$ dependency may be avoidable.

We emphasize that although the sample complexity is exponential in the depth, even complex networks like the QMR-DT network can be shown to have very small maximal depths. In fact, the vast majority of the parameters of the QMR-DT network can be learned with no adjustment at all (i.e., at a depth of 0).

## 4 Experiments

Our first set of experiments look at parameter recovery in samples drawn from a simple synthetic network with the structure of Figure 1, and compare against the variational EM algorithm of Šingliar & Hauskrecht (2006). This network was chosen because it is the simplest network that requires our method to perform the adjustment procedure to learn some of the parameters.

The comparison is done on a small model to show that that even in this simple case, the variational EM baseline performs poorly. Any larger network could have a subnetwork that looks like the network in Figure 1. In our second set of experiments, we apply our algorithm to the large QMR-DT network and show that our algorithm's performance scales to large models.

### 4.1 Comparison with (Variational) EM

Our method-of-moments algorithm is compared to variational EM on 64 networks with the structure of

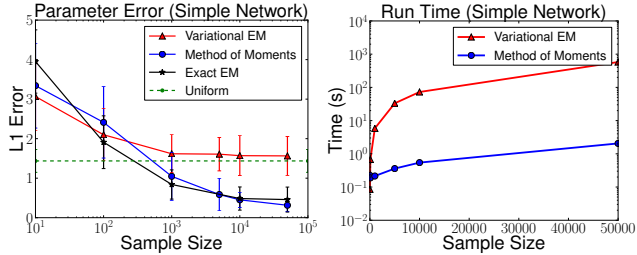

Figure 3: (left) Sum of L1 errors from the true parameters. Error bars show standard deviation from the mean. The dotted line for Uniform denotes the average error from estimating the failures of the noise parent as 1 and failures and priors of all other parents uniformly as 0.5. (right) Run time in seconds of a single run using the network structure from Figure 1 (shown in log scale).

Figure 1 and random parameters. The failure and prior parameters of each network were generated uniformly at random in the range [0.2, 0.8]. The noise probabilities are set to $\nu = 0.01$. For all algorithms, the true structure of the network was provided and only the parameters were left to be estimated. With insufficient data, method-of-moments can estimate parameters outside of the range [0,1]. Any invalid parameters are clipped to lie within $[10^{-6}, 1 - 10^{-6}]$. Since the variational algorithm can become stuck at local maxima, it was seeded with 64 random seeds for each random network and the run that has the best variational lower bound on the likelihood was reported.

Figure 3 shows the L1 error in parameters and run times of the algorithms as a function of the number of samples, averaged over the 64 different networks. Error bars show standard deviation from the mean. The timing test was run on a single machine. Variational EM was run using the authors' C++ implementation of the algorithm[2] and Algorithm 1 was run using a Python implementation. In the large data setting, the method-of-moments algorithm is much faster than variational EM because it only has to iterate through the data once to form empirical estimates of the triplet moments. The variational method requires a pass through the data for every iteration.

In nearly all of the runs, variational EM converges to a set of parameters that effectively assign the children $b$ and $c$ in the network (Figure 1) to one of the two parents $A$ or $B$ by setting the failure probabilities of the other parent to very close to 1. Thus, even though it was provided with the correct structure, the variational EM algorithm effectively pruned out some edges from the network. This bias of the variational EM algorithm towards sparse networks was already noted in Šingliar & Hauskrecht (2006) and appears to be a significant detriment to recovery of the true network parameters.

In addition to the variational EM algorithm, we also show results for EM using exact inference, which is feasible for this simple structure. Exact EM was tested on 16 networks with random parameters and used 4 random initializations, with the run having the best likelihood being reported. These results serve two purposes. First, we want to understand whether the failure of variational EM is due to the error introduced by mean-field inference approximation or due to the fact that EM only reaches a local maxima of the likelihood. The fact that exact EM significantly outperforms variational EM suggests that the problem is with the variational inference. The second purpose is to compare the sample complexity of our method-of-moments approach with a maximum-likelihood based approach. On this small network, the sample complexity of the two approaches appears to be comparable. We emphasize that the exact EM method would be infeasible to run on any reasonably sized network due to the intractability of exact inference in these models.

### 4.2 Synthetic Data from aQMR-DT

We use the Anonymized QMR Knowledge Base[3] which has the same structure as the true network, but the names of the variables have been obscured and the parameters perturbed. To generate the synthetic data, we transform the parameters of the anonymized knowledge base to parameters of a noisy-or Bayesian network using the procedure described in Morris (2001). The disease priors (not given in aQMR-DT) were sampled according to a Zipf law with exponentially more low probability diseases than high probability diseases.

Using Algorithm 1 extended to take advantage of singly-coupled pairs (as described in Section 3.3), we find a schedule with depth 3 that learns all but a single highly connected subnetwork of QMR-DT. This troublesome subnetwork has two parents, each with 61 children, that overlap on all but one child each (similar to Figure 2 but 60 overlapping children instead of 3). It cannot be learned fully using the main algorithm, though it can be learned with the clean up procedure described in Section 3.3.

The pairs method is very useful for decreasing the maximum depth of the network. Figure 4 (right) compares the depths of parameters learned only with the triplet method to those learned using triplets and pairs combined. Using only triplets eventually learns all of the same parameters as using both triplets and pairs, but requires more adjustment steps.

---

[2]We thank the authors of Šingliar & Hauskrecht (2006) for kindly providing their implementation.

[3]The QMR Knowledge Base is provided by University of Pittsburgh through the efforts of Frances Connell, Randolph A. Miller, and Gregory F. Cooper.

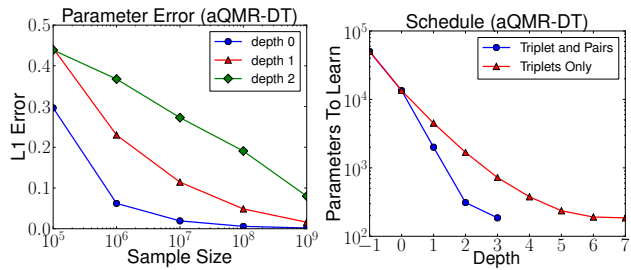

Figure 4: (Left) Mean parameter error as a function of sample size for the failure parameters learned at different depths on the QMR-DT network. Only a small number of failure parameters are learned at depth 3 so it is not included due to its high variance. (Right) Number of parameters (in log scale) left to learn after learning all of the parameters at a given depth, using a schedule that uses both triplets and pairs, compared to a schedule that only uses triplets. At the outset of the algorithm (depth=-1), all of the parameters remain to be learned. The remaining parameters belong to a single subnetwork in the QMR-DT graph that we can learn with the clean up step.

Figure 4 (left) shows the average L1 error for parameters learned as a function of the depth they were learned at. As expected the error compounds with depth, but with sufficiently large samples, all of the errors tend toward zero. Additionally, as shown in Figure 4 (right), the vast majority of the parameters are learned at depth 0 and 1.

Timings were reported using an AMD-based Dell R815 machine with 64 cores and 256GB RAM. First, a valid schedule to learn all of the parameters of the aQMR-DT network (except the subnetwork described above) was found using Algorithm 1 extended to use pairs. Finding a schedule took 4.5 hours using 32 processors in parallel. Once the schedule is determined, the learning procedure only requires sufficient statistics in the form of the joint distributions of the triplets and pairs and single variables present in the schedule (36,506 triplets, 7,682 pairs and 4,013 singles). The network was sampled and sufficient statistics were computed from each sample. Updating the sufficient statistics took approximately $2.5 \cdot 10^{-4}$ seconds per sample and can be trivially parallelized. Solving for the network parameters using the sufficient statistics takes under 3 minutes with no parallelization at all.

## 5 Discussion

We presented a method-of-moments approach to learning the parameters of bipartite noisy-or Bayesian networks of known structure and sufficient sparsity, using unlabeled training data that only needs to observe the bottom layer's variables. The method is fast, has theoretical guarantees, and compares favorably to existing variational methods of parameter learning. We show that using this method we can learn almost all of the parameters of the QMR-DT Bayesian network and provide local identifiability results and a method that suggests the remaining parameters can be estimated efficiently as well.

The main algorithm presented in this paper uses third-order moments, but only recovers parameters of a bipartite noisy-or network for a restricted family of network structures. The clean up algorithm can recover all locally identifiable network structures, including fully connected networks, but requires grid searches for parameters that can be exponential in the number of parents. This leaves open the question of whether there are efficient algorithms for recovering a more expansive family of network structures than those covered by the main algorithm.

Provably learning the *structure* of the noisy-or network as well as its parameters from data is more difficult because of identifiability problems. For example, one can show that third-order moments are insufficient for determining the number of hidden variables. We consider this an open problem for further work. Also, in most real-world applications involving expert systems for diagnosis, the hidden variables are not marginally independent (e.g., having diabetes increases the risk of hypertension). It is possible that the techniques described here can be extended to allow for dependencies between the hidden variables.

Another important direction is to attempt to generalize the learning algorithms beyond noisy-or networks of binary variables. The noisy-or distribution is special because adding parents can only *decrease* the negative moments (Eq. 3), and its factorization allows for the effect of individual parents to be isolated. Moreover, since the noisy-or parameterization has a single parameter per hidden variable and observed variable, it is possible to learn part of the model and then hope to adjust the remaining moments (a more general distribution with the same property is the logistic function). New techniques will likely need to be developed to enable learning of arbitrary discrete-valued Bayesian networks with hidden values.

### Acknowledgements

We thank Sanjeev Arora, Rong Ge, and Ankur Moitra for early discussions on this work. Research supported in part by a Google Faculty Research Award, CIMIT award 12-1262, grant UL1 TR000038 from NCATS, and by an NSERC Postgraduate Scholarship.